\documentclass[runningheads]{llncs}
\usepackage{amsmath}
\usepackage{amssymb}
\usepackage{booktabs} 
\usepackage{wasysym}
\usepackage{graphicx}
\usepackage{url}
\usepackage{tikz}
\usetikzlibrary{arrows, patterns,positioning}
\usetikzlibrary{calc}
\usetikzlibrary{shapes.misc}

\usepackage{pgfplots}

\newlength{\hatchspread}
\newlength{\hatchthickness}
\newlength{\hatchshift}
\newcommand{\hatchcolor}{}
\tikzset{hatchspread/.code={\setlength{\hatchspread}{#1}},
         hatchthickness/.code={\setlength{\hatchthickness}{#1}},
         hatchshift/.code={\setlength{\hatchshift}{#1}},
         hatchcolor/.code={\renewcommand{\hatchcolor}{#1}}}
\tikzset{hatchspread=3pt,
         hatchthickness=0.4pt,
         hatchshift=0pt,
         hatchcolor=black}
\pgfdeclarepatternformonly[\hatchspread,\hatchthickness,\hatchshift,\hatchcolor]
   {custom north west lines}
   {\pgfqpoint{\dimexpr-2\hatchthickness}{\dimexpr-2\hatchthickness}}
   {\pgfqpoint{\dimexpr\hatchspread+2\hatchthickness}{\dimexpr\hatchspread+2\hatchthickness}}
   {\pgfqpoint{\dimexpr\hatchspread}{\dimexpr\hatchspread}}
   {
    \pgfsetlinewidth{\hatchthickness}
    \pgfpathmoveto{\pgfqpoint{0pt}{\dimexpr\hatchspread+\hatchshift}}
    \pgfpathlineto{\pgfqpoint{\dimexpr\hatchspread+0.15pt+\hatchshift}{-0.15pt}}
    \ifdim \hatchshift > 0pt
      \pgfpathmoveto{\pgfqpoint{0pt}{\hatchshift}}
      \pgfpathlineto{\pgfqpoint{\dimexpr0.15pt+\hatchshift}{-0.15pt}}
    \fi
    \pgfsetstrokecolor{\hatchcolor}
    \pgfusepath{stroke}
   }

\usepackage{pgfplots}
\usepackage{xcolor}

\begin{document}
\title{Reactive Control Meets Runtime Verification:\\A Case Study of Navigation}
%
%
\author{Dogan Ulus and Calin Belta}
\authorrunning{D. Ulus and C. Belta}
%
\institute{Boston University\\Boston, MA, USA}
\maketitle              
\begin{abstract}
This paper presents an application of specification based runtime verification techniques to control mobile robots in a reactive manner.
In our case study, we develop a layered control architecture where runtime monitors constructed from formal specifications are embedded into the navigation stack.
We use temporal logic and regular expressions to describe safety requirements and mission specifications, respectively.
An immediate benefit of our approach is that it leverages simple requirements and objectives of traditional control applications to more complex specifications in a non-intrusive and compositional way.
Finally, we demonstrate a simulation of robots controlled by the proposed architecture and we discuss further extensions of our approach.

\end{abstract}
\section{Introduction}
Mobile robots are designed to work either in static and fully predictable environments such as automated warehouses or in open, partially unknown, and constantly changing environments such as road traffic.
Classical deliberative control (complete planning before execution) often work well for the former case while being inadequate or very inefficient for the latter.
Alternatively, in reactive control, robots continuously observe the environment and thus are able to react and adapt to previously unknown circumstances.
%
%
Runtime verification (RV), on the other hand, is a branch of formal methods that deals with checking correctness temporal sequences against high-level specifications~\cite{bartocci2018specification,havelundruntime,sanchez2018survey}.
A common point between reactive control and runtime verification is that they both trade the completeness guarantees of deliberate control and model checking for online computation, practicality, and scalability.
Following this synergy and growing interest in robotics using formal specifications, we think runtime verification techniques can be very useful to raise the level of abstraction and assurance for reactive controllers in robotic applications.

In this paper, we explore the combination of reactive control and runtime verification techniques to construct controllers for mobile robots that satisfy given safety requirements and high-level mission specifications.
To this end, we employ a multi-layered architecture that can be seen in many reactive controllers and enhance each layer with runtime monitors to enforce desired properties on-the-fly.
Compared to existing deliberative control approaches based on game theory~\cite{alur2018compositional,decastro2018collision,kress2018synthesis}, model checking~\cite{chen2012formal,belta2017formal}, and optimization~\cite{ulusoy2014receding,shoukry2016scalable,raman2014model}, our proposal has several advantages summarized as follows.
First, runtime verification techniques require fewer assumptions about the environment and robots as well as scale much better than those exhaustive methods.
%
%
Second, we can use more expressive specification languages for runtime monitor construction.
For example, it is possible to construct efficient runtime monitors from timed, quantitative, and first-order extensions of linear temporal logic (LTL) and regular expressions whereas game-theoretic reactive synthesis is usually limited to a fragment of LTL.
Hence RV techniques cover a larger class of requirements we need in robotic applications.
Third, most runtime verification techniques are compositional and it is easier to build modular robotic systems using them.
All these advantages would suggest a wider applicability of runtime verification techniques in robotics.

%

\begin{figure}[t]
\centering
\tikzset{
>=stealth',
punkt/.style={rectangle, rounded corners, draw=black, very thick, text width=6.5em, minimum height=2em, text centered},
pil/.style={->,thick,shorten <=2pt,shorten >=2pt,}
}
\resizebox{.8\textwidth}{!}{
\begin{tikzpicture}[node distance=1.5cm, auto, scale=0.45]
\node[punkt] (mission) {Monitor Mission};
\node[punkt, right of=mission, xshift=2cm] (select2) {Select Goal};

\node[punkt, below of=mission] (req1) {Monitor Requirements};
\node[punkt, left of=req1, xshift=-2cm] (gen1) {Generate Routes};
\node[punkt, right of=req1, xshift=2cm] (select1) {Select Sub-Goal};

\node[punkt, below of=req1] (req0) {Monitor Requirements};
\node[punkt, left of=req0, xshift=-2cm] (gen0) {Generate Trajectories};
\node[punkt, right of=req0, xshift=2cm] (select0) {Select Controls};

\draw[punkt,dashed] ($(mission.north west)+(-0.5,0.5)$) rectangle ($(req0.south east)+(0.5,-0.5)$);

\draw[punkt,dashed] ($(gen1.north west)+(-0.5,0.5)$) rectangle ($(select0.south east)+(0.5,-0.5)$);

\node[anchor=south] (rv) at ($(mission.north)+(0,0.5)$) {Runtime Verification};
\node[anchor=south, rotate=90] (rv) at ($0.5*(gen1.west)+0.5*(gen0.west)-(0.5,0)$) {Reactive Control};

\draw[pil] (gen1) -- (req1);
\draw[pil] (req1) -- (select1);
\draw[pil] (gen0) -- (req0);
\draw[pil] (req0) -- (select0);

\draw[pil] (mission) -- (select2);

\draw[pil] (select2) -- (select1);
\draw[pil] (select1) -- (select0);
\draw[pil] (select0) -- ($(select0.east)+(3,0)$);

\filldraw ($(select0.east)+(5,0)$) circle [radius=1];
\filldraw ($(select0.east)+(4,0)$) circle [x radius=0.25, y radius=1];
\filldraw ($(select0.east)+(6,0)$) circle [x radius=0.25, y radius=1];

\node[anchor=south] (robot) at ($(select0.east)+(5,1)$) {Robot};
\end{tikzpicture}
}
\caption{The navigation stack used in the case study}
\label{fig:arch}
\vspace{-0.3cm}
\end{figure}
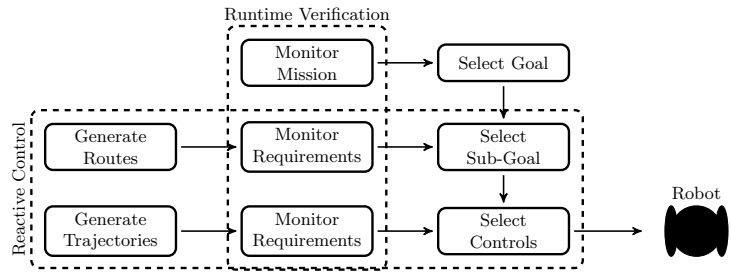

We depict our navigation architecture that contains several components from reactive control and runtime verification domains in Figure~\ref{fig:arch}. 
At the bottom layer of the architecture, we employ limited trajectory search to devise the short-time motion of the robot.
Runtime monitors are embedded to check generated (candidate) trajectories and enforce low-level safety properties such as collision avoidance and one-way regulations.
The middle layer addresses the shortcomings of short-horizon trajectories by searching a route over a connectivity graph of the environment.
Mid-level safety properties for the graph traversal (e.g avoiding specific areas) are similarly enforced using runtime monitors in this layer.
%
%
Once undesired trajectories and routes are filtered out, we use a number of features and heuristics to select the best one among the remaining.
Repeating these procedures in real-time produces a safe motion for the robot to reach a specific (goal) location relative to trajectory/route generation specifics.
Finally, the top layer is designated for high-level mission control that enforces the correct order of locations to be visited.
In our study, we similarly employ runtime monitors constructed from mission specifications for the mission control.

The structure of this paper is as follows. 
First we describe our setup for the case study in Section 2.
We then explain the use of runtime monitors to produce safe motion in Section 3 and to guide missions in Section 4.
The paper is concluded in Section 5 by a discussion of simulation results and future directions.

\section{Environment, Robots, and Specifications}
\begin{figure}[b!]
\centering
\resizebox{.45\textwidth}{!}{%
\begin{tikzpicture}

\tikzset{cross/.style={cross out, draw=red, line width=25pt, minimum size=12*(#1-\pgflinewidth), inner sep=0pt, outer sep=0pt},
cross/.default={30pt}}

\filldraw[fill=yellow] (10.0, 22.3) -- (16.5, 28.8) -- (28.5, 28.8) -- (35.0,22.3) -- cycle;
\filldraw[fill=yellow] (10.0, 35.3) -- (16.5, 28.8) -- (28.5, 28.8) -- (35.0,35.3)  -- cycle;
\filldraw[fill=yellow] (58.0, 22.3) -- (64.5, 28.8) -- (85.9, 28.8) -- (92.4,22.3) -- cycle;
\filldraw[fill=yellow] (58.0, 35.3)  -- (64.5, 28.8) -- (85.9, 28.8) -- (92.4,35.3) -- (75,35.3) -- (72.5,32.3) -- (70,35.3) -- cycle;

\draw[->, line width=20] (16.5, 25.15) -- (28.5, 25.15);
\draw[->, line width=20] (28.5, 31.65) -- (16.5, 31.65);

\draw[->, line width=20] (64.5, 25.15) -- (83.5, 25.15);
\draw[->, line width=20] (83.5, 31.65) -- (64.5, 31.65);

\draw[line width=20] (0,0) rectangle (102.4,57.6);
\draw[line width=20] (0, 22.3) -- (5.0, 22.3);
\draw[line width=20] (22.5, 22.3) -- (22.5, 10.0);
\draw[line width=20] (22.5, 5.0) -- (22.5, 0);
\draw[line width=20] (10.0, 22.3) -- (35.0, 22.3) -- (40,17.3) -- (40,0);

\draw[line width=20] (0, 35.3) -- (5.0, 35.3);
\draw[line width=20] (10.0, 35.3) -- (35.0, 35.3) -- (40.0, 40.3) -- (40.0,57.6);
\draw[line width=20] (53.0, 0.0) -- (53.0, 5);
\draw[line width=20] (53.0, 10.0) -- (53.0, 17.3) -- (58.0, 22.3) -- (92.4, 22.3); 
\draw[line width=20] (97.4, 22.3) -- (102.4, 22.3);
\draw[line width=20] (53.0, 57.6) -- (53.0, 40.3) -- (58.0, 35.3) -- (70.0, 35.3);
\draw[line width=20] (75.0, 35.3) -- (92.4, 35.3);
\draw[line width=20] (97.4, 35.3) -- (102.4, 35.3);
\draw[line width=20] (80.0, 57.6) -- (80.0, 35.3);

\draw (7.5,35.3) node[cross] {};
\draw (7.5,22.5) node[cross] {};
\draw (22.5,7.5) node[cross] {};
\draw (53.0,7.5) node[cross] {};
\draw (72.5,35.3) node[cross] {};
\draw (94.9,35.3) node[cross] {};
\draw (94.9,22.5) node[cross] {};

\draw (61.25,31.65) node[cross] {};
\draw (61.25,25.15) node[cross] {};
\draw (89.15,31.65) node[cross] {};
\draw (89.15,25.15) node[cross] {};
\draw (13.25,31.65) node[cross] {};
\draw (13.25,25.15) node[cross] {};
\draw (31.75,31.65) node[cross] {};
\draw (31.75,25.15) node[cross] {};


\node[font=\Large,scale=10] (r1) at  (5.0, 52.6) {$R1$};
\node[font=\Large,scale=10] (r2) at  (58.0,52.6) {$R2$};
\node[font=\Large,scale=10] (r3) at  (97.4, 52.6) {$R3$};
\node[font=\Large,scale=10] (r4) at  (97.4, 5.0) {$R6$};
\node[font=\Large,scale=10] (r3) at  (35.0, 5.0) {$R5$};
\node[font=\Large,scale=10] (r4) at  (5.0, 5.0) {$R4$};

\node[anchor=south, scale=10, red] (01text) at (7.5,36.3){\large\texttt{D1}};
\node[anchor=south, scale=10, red] (02text) at (72.5,36.3){\large\texttt{D2}};
\node[anchor=south, scale=10, red] (02text) at (94.9,36.3){\large\texttt{D3}};
\node[anchor=north, scale=10, red] (01text) at (7.5, 21.5){\large\texttt{D4}};
\node[anchor=west, scale=10, red] (06text) at (23.5,7.5){\large\texttt{D5}};
\node[anchor=west, scale=10, red] (06text) at (54.0,7.5){\large\texttt{D6A}};
\node[anchor=north, scale=10, red] (02text) at (94.9, 21.5){\large\texttt{D6B}};

\fill[green] (80.0, 10.0) circle (5.0);
\fill[green] (70.0, 50.0) circle (5.0);
\fill[green] (32.0, 15.0) circle (5.0);
\fill[green] (30.0, 45.0) circle (5.0);

\node[scale=20, black] (Atext) at (80.0, 10.0){\texttt{A}};
\node[scale=20, black] (Btext) at (70.0, 50.0){\texttt{B}};
\node[scale=20, black] (Ctext) at (32.0, 15.0){\texttt{C}};
\node[scale=20, black] (Dtext) at (30.0, 45.0){\texttt{D}};

\end{tikzpicture}%
}
\quad\quad
\resizebox{.45\textwidth}{!}{%
\begin{tikzpicture}

\tikzset{cross/.style={circle, fill=red, line width=2pt, minimum size=0.2*(#1-\pgflinewidth), inner sep=3pt, outer sep=0pt, scale=1},
cross/.default={30pt}}

\node[cross] (01) at (0.75,5.53){};
\node[cross] (02) at (0.75,1.25){};
\node[cross] (03) at (2.25,0.75){};
\node[cross] (04) at (4.65,1.75){};

\node[cross] (05) at (7.25,5.53){};
\node[cross] (06) at (9.49,5.53){};
\node[cross] (07) at (9.49,1.25){};

\node[cross] (08) at (6.125,4.165){};
\node[cross] (09) at (6.125,2.515){};
\node[cross] (10) at (8.915,4.165){};
\node[cross] (11) at (8.915,2.515){};
\node[cross] (12) at (1.325,4.165){};
\node[cross] (13) at (1.325,2.515){};
\node[cross] (14) at (3.175,4.165){};
\node[cross] (15) at (3.175,2.515){};

\node[cross,fill=green] (d) at (2.25,5.53){};
\node[cross,fill=green] (b) at (4.75,5.53){};
\node[cross,fill=green] (a) at (7.25,1.5){};
\node[cross,fill=green] (c) at (3.75,0.75){};


\draw[->,> = latex, ultra thick] (14) to[bend right] (12);
\draw[->,> = latex, ultra thick] (12) to[bend right] (01);
\draw[->,> = latex, ultra thick] (12) to[bend right] (13);
\draw[<->,> = latex, ultra thick] (01) to[bend right] (02);

\draw[<->,> = latex, ultra thick] (02) to[bend right] (03);
\draw[->,> = latex, ultra thick] (02) to[bend right] (13);
\draw[->,> = latex, ultra thick] (13) to[bend right] (15);
\draw[->,> = latex, ultra thick] (15) to[bend left] (09);
\draw[->,> = latex, ultra thick] (09) to[bend right] (11);
\draw[->,> = latex, ultra thick] (11) to[bend right] (07);
\draw[<->,> = latex, ultra thick] (04) to[bend right] (07);
\draw[->,> = latex, ultra thick] (04) to[bend right, out=15] (14);
\draw[->,> = latex, ultra thick] (04) to[bend right] (09);
\draw[->,> = latex, ultra thick] (15) to[bend right] (14);
\draw[->,> = latex, ultra thick] (08) to[bend left] (14);
\draw[->,> = latex, ultra thick] (10) to[bend right] (08);
\draw[->,> = latex, ultra thick] (11) to[bend right] (10);
\draw[->,> = latex, ultra thick] (08) to[bend right] (09);
\draw[<->,> = latex, ultra thick] (07) to[bend right] (06);
\draw[->,> = latex, ultra thick] (06) to[bend right] (10);
\draw[->,> = latex, ultra thick] (11) to[bend right] (06);
\draw[->,> = latex, ultra thick] (12) to[bend right] (02);
\draw[->,> = latex, ultra thick] (10) to[bend right] (05);
\draw[->,> = latex, ultra thick] (05) to[bend right] (08);
\draw[->,> = latex, ultra thick] (08) to[bend right, in=165] (04);
\draw[->,> = latex, ultra thick] (01) to[bend right] (13);
\draw[->,> = latex, ultra thick] (07) to[bend right] (10);
\draw[->,> = latex, ultra thick] (15) to[bend right] (04);

\draw[<->,> = latex, ultra thick] (01) to[bend left] (d);
\draw[<->,> = latex, ultra thick] (05) to[bend right] (b);
\draw[<->,> = latex, ultra thick] (07) to[bend right] (a);
\draw[<->,> = latex, ultra thick] (04) to[bend right] (a);
\draw[<->,> = latex, ultra thick] (03) to[bend left] (c);


\node[anchor=north east] (04text) at (4.65,1.75){\Large\texttt{D6A}};
\node[anchor=north] (07text) at (9.49,0.95){\Large\texttt{D6B}};
\node[anchor=south] (06text) at (9.49,5.73){\Large\texttt{D3}};
\node[anchor=south] (06text) at (7.25,5.73){\Large\texttt{D2}};
\node[anchor=south] (06text) at (0.75,5.73){\Large\texttt{D1}};
\node[anchor=north east] (06text) at (0.75,1.25){\Large\texttt{D4}};
\node[anchor=south] (06text) at (2.25,0.95){\Large\texttt{D5}};

\node[anchor=south] (d) at (2.25,5.73){\Large\texttt{D}};
\node[anchor=south] (b) at (4.75,5.73){\Large\texttt{B}};
\node[anchor=north west] (a) at (7.35,1.5){\Large\texttt{A}};
\node[anchor=west] (c) at (3.85,0.75){\Large\texttt{C}};
\end{tikzpicture}
}
\caption{Environment maps: Geometric on the left and topological on the right}
\label{fig:env}
\end{figure}
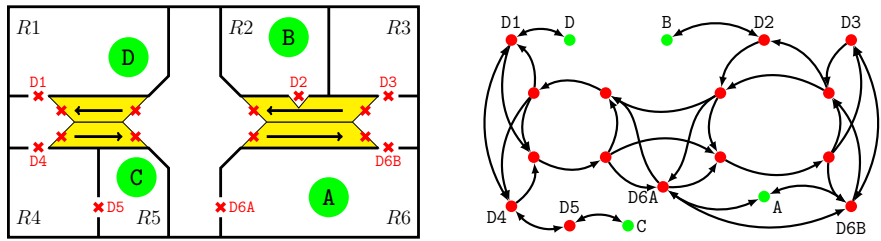

For our case study, we will work on a relatively complex 2D environment designed to give a representative view of real challenges without introducing too much detail.
Depicted on the left of Figure~\ref{fig:env}, our environment represents an office space with rooms (R1-R6), narrow passages (such as doors D1-D6), named locations (A-D), and some regulations at certain regions (one-way regions) including other (possibly uncontrolled) agents.
We use a unicycle velocity-controlled model for the robot dynamics where the state space is defined by robot's position $(x,y)$ and orientation $\theta$, and controlled by forward and angular velocity commands $u = (v, \omega)$.

It is of critical importance that the complexity of the environment determines the complexity of specifications and monitoring.
For a static environment (that is to say, nothing changes outside of our control), we do not need any runtime monitoring at all. 
This is obviously a very strong assumption for many cases.
On the other hand, if dynamic obstacles (such other agents) exist in the environment, we have to at least add a basic monitoring mechanism that checks simple propositions ---will the robot collide with anything soon or did the robot reach its goal?
Moreover, if we have more complex regulations and tasks to complete in the environment, rich specification languages to describe them becomes a preferable option. 
Therefore, our robots are assigned to perform complex navigation missions, specified by regular expressions, while avoiding static and dynamic obstacles as well as satisfying desired properties and regulations, specified by temporal logic formulas.

Finally we explain our use of environment maps depicted in Figure~\ref{fig:env}. 
In the real world, robots sense their immediate vicinity through their various sensors, which we do not model in this case study.
Instead we assume that the locations of nearby static and dynamic obstacles are available to the robot at any time. 
Our control strategy does not depend on the full real-time geometric knowledge of the environment nor its naive discretization (grid-world).
On the other hand, we heavily use the connectivity graph of the environment depicted in the right of Figure~\ref{fig:env}.
Using such topological maps together with local motion planning are much more scalable and reliable than directly planning a full trajectory on a geometric map~\cite{blochliger2018topomap}.

\section{Search for Safe Motion}
In this paper, we use the past fragment of linear temporal logic to specify safety properties due to the causality of the formalism and efficiency of constructed monitors. 
The semantics of past temporal operators \texttt{always}, \texttt{once}, \texttt{since}, and \texttt{previously}, their extensions, and monitor constructions from these specifications can be found in the papers~\cite{havelund2004monitoring,monitor-bdd,online-mtl}.

We here show an application of runtime monitors to enforce the desired safe behavior inside trajectory and route search algorithms.
The general procedure can be summarized in three steps: (1) Generating a number of alternative behaviors (trajectories or routes), (2) checking unsafe/undesired behaviors using runtime monitors, and (3) selecting the best remaining (thus safe) behaviors according to a predefined set of heuristics.
Importantly, the extent of these search processes is limited due to available computational resources as well as that long-term complete plans may become invalid very quickly in dynamic and uncertain environments.
In the following, we give more details about search procedures and actual properties used in the case study.

%

%

%

%
%
%

%

%
%
%
%
%
\subsubsection*{Trajectory Search}
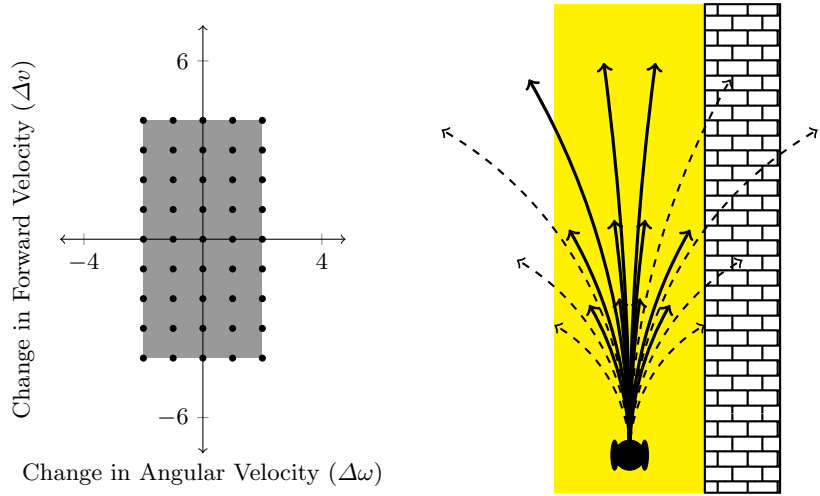
\begin{figure}[t]
\centering
\begin{tikzpicture}[scale=1]
\begin{axis}[
    unit vector ratio*=1 1 1,
    axis lines=center, axis on top,
    axis line style={<->},
    enlargelimits = true,
    xlabel = {Change in Angular Velocity $(\Delta\omega)$},
    ylabel = {Change in Forward Velocity $(\Delta v)$},
    x label style={at={(axis description cs:0.5,0)},anchor=north},
    y label style={at={(axis description cs:-0.2,0.5)},anchor=north, rotate=90},
    ytick={-6,0,6},
    xtick={-4,0,4},
    xmin=-4, xmax=4, ymin=-6, ymax=6]

    \fill[->, thick, gray!80!white] (axis cs: -2, -4) rectangle (axis cs: 2, 4);
    
    \foreach \x in {-2,...,2}
    \foreach \y in {-4,...,4}{
        \edef\temp{\noexpand\filldraw (axis cs: \x,\y) circle [radius=10];}
        \temp
    }

\end{axis}
\end{tikzpicture}%
\quad\quad
\begin{tikzpicture}[scale=1]

\fill[yellow] (-1, -0.5) rectangle (1, 6);
\filldraw[->, thick, pattern=bricks] (1, -0.5) rectangle (2, 6);

\filldraw (0,0) circle [radius=0.2];
\filldraw (0.2,0) circle [x radius=0.05, y radius=0.2];
\filldraw (-0.2,0) circle [x radius=0.05, y radius=0.2];



\draw[->, thick, dashed] (0,0) arc (0:60:2);
\draw[->, very thick] (0,0) arc (0:30:4);
\draw[->, very thick] (0,0) arc (0:7.5:16);
\draw[->, thick, dashed] (0,0) arc (180:120:2);
\draw[->, very thick] (0,0) arc (180:150:4);
\draw[->, very thick] (0,0) arc (180:172.5:16);

\draw[->, thick, dashed] (0,0) arc (0:60:3);
\draw[->, very thick] (0,0) arc (0:30:6);
\draw[->, very thick] (0,0) arc (0:7.5:24);
\draw[->, thick, dashed] (0,0) arc (180:120:3);
\draw[->, very thick] (0,0) arc (180:150:6);
\draw[->, very thick] (0,0) arc (180:172.5:24);

\draw[->, thick, dashed] (0,0) arc (0:60:5);
\draw[->, very thick] (0,0) arc (0:30:10);
\draw[->, very thick] (0,0) arc (0:7.5:40);
\draw[->, thick, dashed] (0,0) arc (180:120:5);
\draw[->, thick, dashed] (0,0) arc (180:150:10);
\draw[->, very thick] (0,0) arc (180:172.5:40);

\end{tikzpicture}
\quad\quad
\caption{(Left) A finite set of admissible velocity commands for the next time step relative to the current velocity. The search space, depicted in gray, is constrained by maximum allowed accelerations of the robot. (Right) Future trajectories of the robot simulated for each admissible velocity command. Dashed trajectories contain a violation in specification so commands that leads to these trajectories are discarded.}
\label{fig:dwa}
\end{figure}

Dynamic window approach (DWA)~\cite{fox1997dynamic} is a well-known collision avoidance and local motion planning algorithm that uses search procedures to find control actions (velocity commands) while considering robot's dynamics. 
The search space of DWA is limited by maximum acceleration available to the robot as depicted on the left of Figure~\ref{fig:dwa} and the algorithm samples a set of control actions from this space typically over a grid of some resolution. 
Then it calculates the future trajectories (over a limited time horizon) of each alternative action as illustrated on the right of the figure.

Originally being a collision avoidance algorithm, the only safety requirement over these trajectories considered in DWA is never getting dangerously close to obstacles, which is usually hard-coded into the algorithm.
On the other hand, we are interested in checking such requirements using runtime monitors so that we can extend the approach for any temporal logic formula.
We start our case study by expressing the collision avoidance requirement in temporal logic as follows. 
\begin{equation}
\texttt{never( dangerously\_close(obstacles))} \tag{CA}
\end{equation}
where \texttt{dangerously\_close} is a predicate that computes whether any intersection occurs between obstacles and robot's extended footprint.
%
%
%
%
%

In this case study, besides collision avoidance, we also want our robot to obey one-way regulations of the environment, which state that robots have to move in a single direction inside certain regions.
The direction of one-way regions is either west or east in our environment. 
We call these regions westways and eastways accordingly.
Predicates \texttt{inside\_westway} and \texttt{inside\_eastway} check whether the robot in these regions.
Moreover, we define some auxiliary formulas to detect whether the robot just entered a one-way region such that
\begin{align*}
\texttt{entered\_eastway} & \texttt{ : inside\_eastway and not previously inside\_eastway}\\
\texttt{entered\_westway} & \texttt{ : inside\_westway and not previously inside\_westway}
\end{align*}
The correct direction in a one-way region is checked by predicates
\begin{align*}
\texttt{going\_east} & \texttt{ : 75 < orientation <  105 }\\
\texttt{going\_west} & \texttt{ : 255 < orientation < 285 }
\end{align*}
where \texttt{orientation} denotes the current orientation of the robot in degrees (\texttt{0-360}) and the zero degree denote the north.
%
%
%
Finally we write our safety properties for each type of one-way regions as follows.
\begin{align*}
\texttt{inside\_eastway implies (going\_east since entered\_eastway} \tag{OW-E})\\
\texttt{inside\_westway implies (going\_west since entered\_westway} \tag{OW-W})
\end{align*}
However, notice that having many one-way specifications can lead to a linear growth in formula and monitor size.
A better way to handle these requirements is to use parametric specifications where the direction of one-way region is considered as a parameter in the formula.
The conjuction of formulas (OW-W) and (OW-E) can be expressed more concisely as follows.
%
%
%
%
\begin{align*}
\texttt{forall X. inside(X) implies (going(X) since entered(X)}\tag{OW})
\end{align*}
where $\texttt{X} \in \{\texttt{eastway}, \texttt{westway}\}$.
Monitoring such first-order temporal formulas is solved very efficiently in~\cite{monitor-bdd}.
Finally, we construct our runtime monitor to check the conjunction of (CA) and (OW) requirements over trajectories. 
Control actions that produces violating trajectories are discarded before the selection phase.
This ensures the safety of selected control action if there exists one in alternatives otherwise we apply full break.

The last piece of trajectory search is to select the best one among safe trajectories according to a weighted sum of some predefined heuristics, namely final speed of the trajectory (higher is better), final-distance-to-goal (lower is better), minimum-distance-to-obstacles (higher is better).
In the case study, the actual values of weights are found empirically.
\subsubsection*{Route Search} 
A well-known problem of local trajectory search algorithms like DWA is that they may cause a robot to be stuck at a corner if the navigable space is highly non-convex. 
This problem can be mitigated by stacking an additional layer of control to steer the robot over key locations (such as doors, intersections, etc.) of the environment .
Given a connectivity graph of these locations, we can search a route from the current location to the actual goal location and each node on the route is passed to the lower layer as a (sub) goal.
In search of a suitable route, we need to take into account some extra requirements. 
First, it is often desired the chosen route do not contain any loops, thus be a simple path over the graph.
We can specify the simple path property as follows.
\begin{align*}
\texttt{forall X. visit(X) -> not once visit(X)}\tag{SP}
\end{align*}
The SP property is usually embedded into graph search algorithms via marking visited nodes; therefore, we may or may not use an external monitor depending on our route generation technique.
On the other hand, external runtime monitors are desirable to enforce application-specific properties as in trajectory search rather than generating a new graph search algorithm for each and every of them. 
For example, consider a property such that the robot never uses  the door \texttt{D6A} when going from the location \texttt{D} to \texttt{A}, which can be expressed as follows.
\begin{align*}
\texttt{(visit(A) \&\& once visit(D)) -> (!visit(D6A) since visit(D))}\tag{ND}
\end{align*}
We then construct a runtime monitor from the property ND (and SP if needed) to check routes generated over the graph.
In particular, we use an off-the-shelf implementation of the shortest path algorithm~\cite{yen1971finding} that generates simple paths starting from the shortest one in this paper.
Sequentially checking these paths using runtime monitors constructed from temporal logic formulas~\cite{havelund2004monitoring,online-mtl} ensures that the we select the shortest route that satisfies specified properties and then  we can update the route of the robot accordingly.

\section{Navigate by Regular Expressions}
In this section, we use regular expressions to specify complex navigation missions and guide the mission execution via runtime monitors constructed from the specification.
Navigation missions describe the desired behavior of the robot over a set of observations and regular operations of sequential composition (\texttt{;}), alternative choice (\texttt{|}), and repetition (\texttt{*}) are used to express the ordering between these observations.
%
%
For example, a robot is said to reach a region \texttt{A} when it was outside for a while and then entered the region \texttt{A}. 
We can specify such a behavior using regular expressions as follows:
\begin{equation*}
    \texttt{reach(A) = (outside(A)){*}; inside(A)}
\end{equation*}
where atomic propositions \texttt{inside(A)} and \texttt{outside(A)} check whether the robot is in the region \texttt{A} or not.
Similarly more complex missions are obtained by composing simple missions as below.
\begin{equation*}
\texttt{mission1 : (reach(C); reach(B)|reach(D); reach(A))*}\tag{M1}
\end{equation*}
which specifies a (robot) behavior to repeatedly visit the regions \texttt{A}, and \texttt{D} with visiting \texttt{B} or \texttt{C} in-between.
From this expression, we construct a runtime monitor~\cite{ulus2018sequential} that associates a Boolean state variable for each proposition and updates them according to previous states and robot's position at each time step.

Recall that the purpose of the mission layer is to guide the mission execution by sending the next goal location to the lower levels.
Therefore, we extend the runtime monitor to output goal locations according to the state of the mission in the following.
For that we augment each state variable corresponding to a proposition with a set of goal locations. 
For example, the proposition \texttt{inside(C)} is augmented with $\{\texttt{B}, \texttt{C}\}$ according to the follow sets (see~\cite{ulus2018sequential}) and \texttt{outside(A)} with the empty set.
Intuitively it means that the robot is ordered to go to the locations \texttt{B} or \texttt{C} when it is outside and go nowhere if it is inside.
The final output of the monitor obtained by taking the union of activated goals as we may have several alternative ways to complete a mission.
Notice that this is essentially due to non-determinism induced by choice and repetition operators of regular expressions.
Finally we note that we randomly select one goal location in this study if the output of the monitor contains more than one location without any further consideration. 

\begin{figure}[t]
\centering
\begin{tabular}{cc}
\texttt{(C;(B|D);A)*} & \texttt{(A;B;C)*}\\
\includegraphics[draft=false, width=.45\textwidth, trim={0 0.8cm 0 0.8cm}, clip]{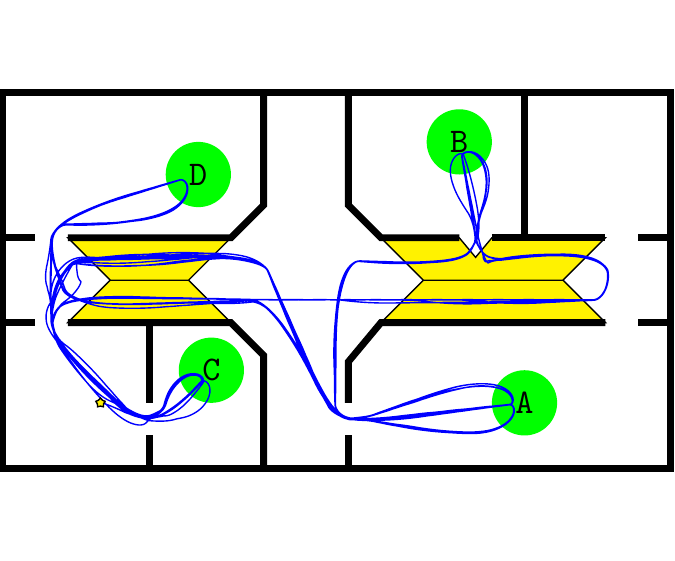}     & \includegraphics[draft=false,width=.45\textwidth, trim={0 0.8cm 0 0.8cm}, clip]{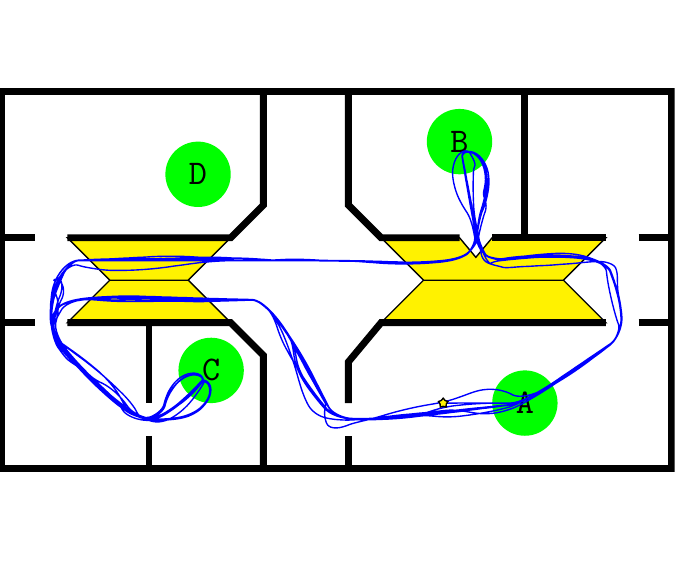} \\
\texttt{A;(D;B;C)*} & \texttt{(A;B;(C|D))*} \\
\includegraphics[draft=false,width=.45\textwidth, trim={0 0.8cm 0 0.8cm}, clip]{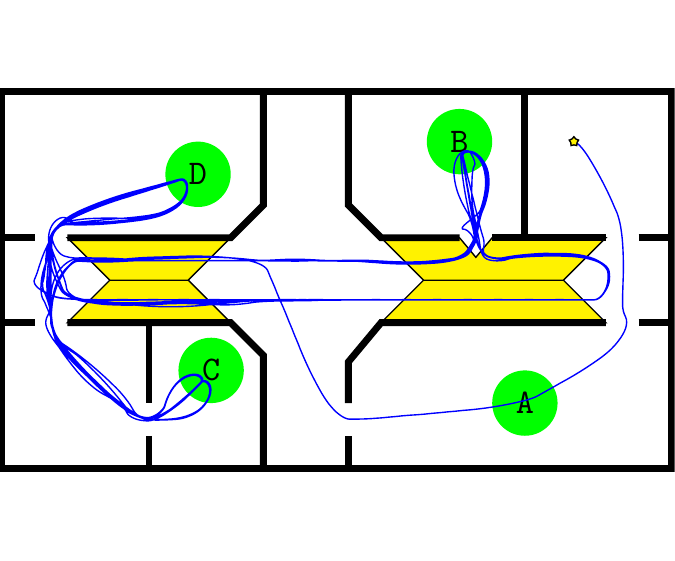}     & \includegraphics[draft=false,width=.45\textwidth, trim={0 0.8cm 0 0.8cm}, clip]{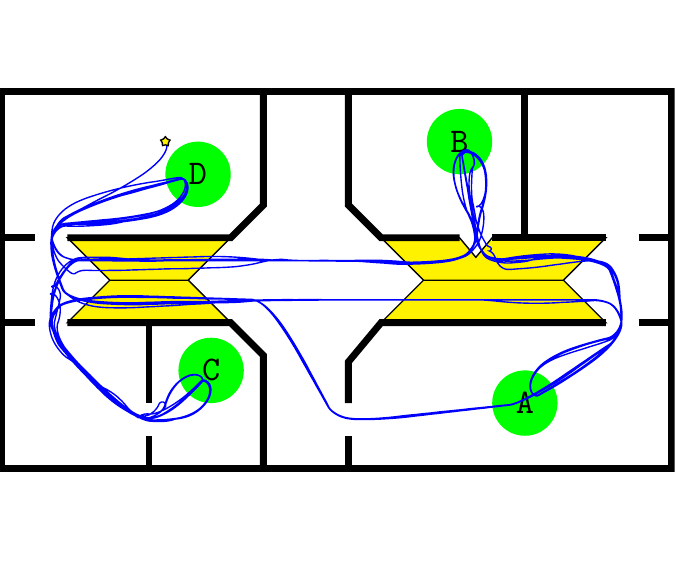}
\end{tabular}

\caption{Trajectories of robots G1-G4 assigned with missions M1-M4, respectively.}
\label{fig:simulation}
\end{figure}

Finally we present our simulation results of four robots G1-G4 operated in the same environment and controlled by the proposed architecture.
We assign the first robot G1 with the mission M1 and the rest G2-G4 with missions M2-M4 below, respectively.
\begin{align*}
\texttt{mission2 : }& \texttt{(reach(A); reach(B); reach(C))*}\tag{M2}\\
\texttt{mission3 : }& \texttt{(reach(A);(reach(D);reach(B); reach(C))*}\tag{M3}\\
\texttt{mission4 : }& \texttt{(reach(A); reach(B); (reach(C)|reach(D))*}\tag{M4}
\end{align*}
In Figure~\ref{fig:simulation}, we separately show the simulated trajectories of the robot for a certain duration that covers several loops as specified in the mission. 
The initial position of the robot is marked by a yellow star.
%
%
Robots get close to each other quite frequently and evading maneuvers cause small variations among loops seen in the figure.
Overall we see that the robots successfully avoid each other and static obstacles and obey regulations of the environment while performing their formally-specified missions over achieving near-optimal trajectories.

\section{Discussion}
In this paper, we presented an example and novel use of provably correct runtime monitors to control a mobile robot subject to complex safety requirements and mission specifications in a dynamic environment.
To this end, we embedded runtime monitors into a layered reactive control architecture together with other simple and scalable components to achieve a navigation solution that does not require strong assumptions. 
Our approach amounts to a more active use of runtime monitors beyond checking assumptions of an offline planner at runtime~\cite{lahijanian2016iterative,ayala2013temporal,desai2017combining}.
We believe the simplicity and breadth of runtime monitors would make them ideal to cover many use cases and increase the level of assurance in robotic applications.
In the future, we would like to explore collective missions for a team of robots as well as some relations among missions (e.g. priority). 
We think the compositionality of runtime monitors used in this study would be important to realize these much needed features.
We finally note that, unlike traditional end-to-end motion planning, we have not been interested in generating a complete trajectory to the destination for the robot in this study.
The main drawback as we see is that accurate long-term plans can be invalidated quickly in a highly dynamic environment. 
Hence there is no point of making such plans for the robot as we need to re-plan anyways, perhaps frequently.
On the other hand, it may be still better to envisage these deliberative techniques in a reactive framework to generate short-term trajectories or build/update the connectivity graph.
In that case, we can use runtime monitors to enforce safety requirements as explained in this paper.

\clearpage

\bibliographystyle{splncs04}
\bibliography{ref}

\end{document}